\pdfoutput=1
\documentclass[11pt]{article}

\usepackage[preprint]{acl}

\usepackage{times}
\usepackage{latexsym}

\usepackage[T1]{fontenc}
\usepackage[utf8]{inputenc}

\usepackage{microtype}

\usepackage{inconsolata}
\usepackage{booktabs}
\usepackage{multirow}
\usepackage{url}
\usepackage{hyperref}
\usepackage{amsmath}
\usepackage{graphicx}
\usepackage{subfigure}
\usepackage{authblk}
\title{Bigger But Not Better: Small Neural Language Models Outperform Large Language Models in Detection of Thought Disorder}

\newcommand{\equalcontrib}{\textsuperscript{*}}
\author{
  Changye Li$^{1}$\equalcontrib \quad Weizhe Xu$^{1}$\equalcontrib \quad Serguei Pakhomov$^{2}$ \\
  Ellen Bradley$^{3}$ \quad Dror Ben-Zeev$^{1}$ \quad Trevor Cohen$^{1}$ \\
  \\
  $^{1}$University of Washington \\ $^{2}$University of Minnesota \\ $^{3}$University of California, San Francisco \\
  \texttt{\{changyel, xuweizhe\}@uw.edu}
}

\begin{document}
\maketitle
\renewcommand{\thefootnote}{\fnsymbol{footnote}}
\footnotetext[1]{Equal contribution}
\renewcommand{\thefootnote}{\arabic{footnote}}
\begin{abstract}

Disorganized thinking is a key diagnostic indicator of schizophrenia-spectrum disorders. Recently, clinical estimates of the severity of disorganized thinking have been shown to correlate with measures of how difficult speech transcripts would be for large language models (LLMs) to predict. However, LLMs' deployment challenges -- including privacy concerns, computational and financial costs, and lack of transparency of training data -- limit their clinical utility. We investigate whether smaller neural language models can serve as effective alternatives for detecting positive formal thought disorder, using the same sliding window based perplexity measurements that proved effective with larger models. Surprisingly, our results show that smaller models are more sensitive to linguistic differences associated with formal thought disorder than their larger counterparts. Detection capability declines beyond a certain model size and context length, challenging the common assumption of ``bigger is better'' for LLM-based applications. Our findings generalize across audio diaries and clinical interview speech samples from individuals with psychotic symptoms, suggesting a promising direction for developing efficient, cost-effective, and privacy-preserving screening tools that can be deployed in both clinical and naturalistic settings.

 %To address these issues we explore the utility of probability estimates from smaller neural language models (LMs) using perplexity as indicators of disorganized thinking. Surprisingly, we find that smaller models align as well as clinical scales as, or better than their larger counterparts. Our findings provide empirical evidence that smaller LMs are more sensitive to linguistic patterns associated with positive formal thought disorder
\end{abstract}

\section{Introduction}

With an estimated prevalence of 15.2 in 100,000 persons \citep{mcgrath2008schizophrenia}, schizophrenia-spectrum disorders (SSDs) are debilitating conditions that can lead to impaired social and occupational functioning, and poor healthcare outcomes including early mortality \citep{laursen2014excess}.  Formal Thought Disorder (FTD) -- a breakdown in the structure of an individual's thinking -- is a diagnostic feature of schizophrenia \citep{kircher2018formal}, and is recognized by observing speech that appears incoherent. Traditional evaluation of FTD relies on clinical interviews and standardized rating scales, which require extensive training and can be time-consuming. Natural language processing (NLP) methods have emerged as promising computational tools for automated evaluation of FTD. These data-driven approaches can systematically analyze linguistic patterns and discourse structure in patients' speech, offering objective quantitative measures of semantic coherence that correspond with clinical estimates of FTD severity \citep[\textit{inter alia}]{Elvevag2007, CORCORAN2020158, xu2021centroid, sarzynska2021detecting, XU2022103998}.

Among the language impairments detectable by NLP methods, FTD represents a particularly complex set of disruptions in thought and speech organization. Patients with FTD exhibit distinct thinking patterns including tangentiality (gradual topic drift) and derailment (completely/partially unrelated thoughts), which may indicate a relative insensitivity to global discourse context \citep{kuperberg2010language1, kuperberg2010language2}. Previous psycholinguistic and neurolinguistic studies in \textit{language comprehension} have shown an impaired ability to use global linguistic context (e.g., information from early in longer sentences) and relatively intact ability for local linguistic context (e.g., information from shorter sentences, local priming) in SSDs \citep{sitnikova2002electrophysiological,swaab2013spared}. With recent advances in neural language models (LMs) in particular, it is now possible to measure dependence upon global and local context in \textit{language production} in SSDs. Recent work \citep{sharpe2024gpt} suggests that disorganized speech can be characterized by comparing the global (probabilities estimated for observed text when including proximal and distal context) and local (probabilities estimated for observed speech when including proximal context only) lexical probabilities retrieved from GPT-3 \citep{brown2020language} when applied to speech samples from SSDs patients and neurotypical demographically-matched controls. A key finding from this research is that models that include longer context appear better-equipped to recognize language from participants with SSDs, in accordance with prior work showing SSDs patients are less influenced by overall sentence context during text comprehension \citep{sitnikova2002electrophysiological}. 

%A better use of the next paragraph here might be to identify the disadvantages of LLMs for this purpose, notably: (1) privacy concerns (whereas smaller models could run on device); (2) cost (whereas smaller models could be locally hosted; (3) transparency, as the training data are not public - sources of bias cannot be investigated, and this bias cannot be mitigated by retraining the model. The next paragraph could perhaps be moved to discussion, as it provides a potential explanation for the findings. 

Commercially-developed large language models (LLMs) such as GPT-3 are pre-trained on large text corpora to enhance their linguistic capabilities. However, their usage and deployment raise concerns in clinical settings. Healthcare applications require stringent privacy protections, yet commercial LLMs mostly operate through cloud-based APIs, requiring the sharing of patient data with third-party commercial services. Even services compliant with the Health Information Portability and Accountability Act (HIPAA) or European counterpart General Data Protection Regulation (GDPR), LLMs can present certain risks to patient privacy especially when used for health-related purposes outside of a healthcare environment \citep{marks2023ai}. In addition to privacy concerns, computational requirements and associated costs of accessing commercial LLMs may also restrict their application in clinical settings. Lastly, the proprietary nature of commercial LLMs and a lack of transparency regarding their training data make it difficult to investigate or mitigate sources of bias.

While these limitations present challenges, there is a pressing need to explore the potential of LLMs in healthcare settings. Healthcare systems continuously face significant workforce shortages, particularly in specialized areas requiring extensive training \citep{thomas2009county, butryn2017shortage}. Current diagnostic and monitoring approaches rely heavily on in-person evaluations, creating bottlenecks in patient care and limiting access to specialized services, especially in resource-constrained settings. LLMs, if properly implemented with appropriate privacy safeguards, could help address these challenges by facilitating the development of technology-assisted diagnostic and monitoring tools, potentially improving the efficiency and accessibility of healthcare services.

Beyond these obstacles, we posit that commercial LLMs' extensive exposure to diverse linguistic patterns drawn from the internet and other sources -- while beneficial for their remarkable text generation capabilities -- may paradoxically reduce their sensitivity to subtle linguistic differences. This hypothesis is supported by recent works suggesting that broad exposure to diverse linguistic data leads LLMs to prioritize general patterns over fine-grained linguistic sensitivity \citep{lee-etal-2024-psycholinguistic, cong2024manner, wilson-etal-2023-abstract}, potentially diminishing their sensitivity to subtle deviations characteristic of FTD. We hypothesize that smaller LMs, such as those in the Pythia suite from EleutherAI \citep{biderman2023pythia}, may exhibit enhanced sensitivity to these linguistic phenomena. These models, ranging from 70M to 12B parameters, are trained on identical public datasets in the same order. As they differ in size only, it is possible to assess the extent to which they respond to nuanced linguistic patterns differently with the constraints in their capacity. In contrast, larger models' potentially excessive capacity to model complex textual relationships may obscure these subtle linguistic markers beneath layers of broader contextual understanding learned from vast amounts of data.

LMs' sensitivity to linguistic manifestations can be measured with perplexity (PPL). PPL is an intrinsic measure used to evaluate the performance of language models on unseen data. The more different the input is from a LM's training data, the ``harder'' it is for the model to predict the next word, resulting in higher PPL. Therefore, it is reasonable to hypothesize that PPL may have some degree of diagnostic utility, as has been documented by prior work using PPL to evaluate cognitive impairment in Dementia of the Alzheimer's Type \citep[\textit{inter alia}]{orimaye2018deep, fritsch2019automatic, cohen-pakhomov-2020-tale, li-etal-2022-gpt, li-etal-2024-big} and psychosis \citep{colla2022semantic, he2024navigating}. 

Building upon \citet{sharpe2024gpt}'s findings with LLMs, our study seeks to assess smaller LMs' sensitivity to linguistic patterns associated with \textit{positive} FTD by analyzing PPLs derived from Pythia models (exemplifying smaller LMs) and LLaMA \citep{dubey2024llama} (exemplifying a LLM) across both monologue and conversational speech samples from individuals with psychotic symptoms and clinically diagnosed SSDs respectively, and evaluating their correlation with the corresponding clinical ratings.

The contributions of this work can be summarized as follows: a) we provide empirical evidence suggesting that smaller LMs are more sensitive to linguistic patterns associated with FTD; b) we demonstrate that the degree of sensitivity starts to decline after models exceed a threshold of a certain number of parameters, suggesting a diminishing relationship between model size and detection capability; and c) the sliding window PPL approach generalizes to both monologue and conversational speech samples of individuals with psychotic symptoms and clinically diagnosed SSDs respectively, suggesting its potential utility for screening and monitoring of SSDs in diverse clinical settings.\footnote{Our code is publicly available on \url{https://github.com/LinguisticAnomalies/small-lm-sliding-windows}}

\section{Related Work}

\subsection{FTD in schizophrenia}
Traditionally, FTD is evaluated through clinically administered rating scales such as the thought and language index \citep{liddle2002thought} or thought and language disorder (TALD) scale \citep{Kircher2014} in research settings, which captures the full variety of FTD phenomenology including subjective experiences. It can also be more conveniently evaluated through self-reporting scales \citep{Barrera2008}. However, there are inherent problems associated with each approach: administering the clinical scales is time-consuming and requires specific training and expertise. Additionally, even when the required expertise is readily available, the clinical assessment only provides intermittent measures during office visits, making it difficult to paint a continuous picture in a more ecological setting. On the other hand, the self-reporting scale lacks objectivity because each patient may have subjective views on the scale's severity settings, and the ability to self-appraise may be impaired in psychosis. These inherent problems and the advancement of computational technologies have inspired the usage of NLP methods to evaluate and quantify the severity of FTD. 

\subsection{Assessing FTD and SSDs with NLP methods}
Advancements in computational systems have introduced innovative methods for automated FTD assessment. A seminal approach by \citet{Foltz1998} utilizes distributional similarity, specifically measuring semantic relatedness between consecutive text segments using latent semantic analysis (LSA) \citep{landauer1997solution} to measure coherence, providing a proxy that was later used to quantify the degree FTD.  This method’s diagnostic utility was demonstrated by \citet{Elvevag2007}, who found significant differences in automated coherence metrics when comparing individuals with schizophrenia to healthy controls, as well as among patients with varying levels of thought disorder. 

Building on this work, a subsequent study integrated LSA-based coherence metrics into a machine learning classifier that accurately predicted psychosis onset in a small sample of at-risk youth, achieving perfect leave-one-out cross-validation accuracy \citep{Bedi2015}. An adapted version maintained 83\% accuracy in predicting psychosis onset in a larger, independent dataset \citep{Corcoran2018}. More recently, neural word embeddings \citep{word2vec}, which represent words as vectors derived from neural networks trained to predict nearby words, have been explored as an alternative to LSA for coherence analysis. Similarity metrics from these embeddings showed promising results in aligning with clinical assessments of thought disorder \citep{just-etal-2019-coherence, Just2020}.

With advances in NLP methods, recent studies have used sentence embeddings from BERT (Bidirectional Encoder Representations from Transformers) \citep{devlin-etal-2019-bert} to identify coherence differences between transcripts from individuals with SSDs and those from healthy controls \citep{Tang2021}. As a transformer-based model, BERT generates context-specific representations of tokens by dynamically incorporating information from surrounding words, unlike LSA or neural word embeddings, which rely on static word representations derived from all of the contexts a word is observed in during training. Prior research also introduced methods for assessing global coherence -- estimating the relationship between a unit of text and the overarching theme of a text -- using these methods to improve coherence evaluation in automatic speech recognition by extracting time series features for machine learning \citep{xu2021centroid, XU2022103998}.

With the emergence of autoregressive LMs, some recent studies \citep{palaniyappanStudyingPsychosisUsing2023, FRADKIN20231013, sharpe2024gpt} have examined the assessment of psychosis using such models to demonstrate that LMs can be utilized with in-silico experimental research to gain better understanding of the linguistic manifestation of FTD. In contrast to BERT, which is a bidirectional model that utilizes tokens on both sides of a target token for prediction, autoregressive LMs are designed to predict only the next token in a sequence. While BERT-derived representations are highly effective for estimating semantic relatedness, autoregressive LMs are specifically optimized for generating coherent and fluent sequences of text, offering potential for developing alternative approaches to FTD evaluation. However, these approaches have primarily relied on such models without exploring how model size and granular context windows affect sensitivity to linguistic manifestations of FTD.

\section{Methods}

\subsection{Data}

\paragraph{AVH Dataset} Speech monologue samples from native English speaking participants who experienced auditory verbal hallucinations (AVH) using a smartphone application were collected during the course of a previous study \citep{mobileRDoC}. Participants experiencing AVH were recruited in-person and online, and prompted to describe their experiences of AVH and anything else they would like to share or think would be helpful for the research team to know. Informed consent from participants was obtained through a rigorous procedure involving triple confirmations from a screening questionnaire. The study was approved by the Institutional Review Boards (IRB) of the University of Washington and Dartmouth College. Two annotators labeled the manual transcripts of the audio recordings for their degree of incoherence based on the TALD scale, using the construct of derailment. The TALD score ranges from 0-4 and represents greater incoherence as the score increases. The inter-rater agreement between annotators was 0.71, as measured by weighted Kappa. This set contained samples with a mean TALD score of 1.18 and a standard deviation of 0.83. We select 310 recordings that: a) have manual transcriptions; and b) are annotated with TALD. The transcript-level demographic information for this dataset is summarized in Table~\ref{tab:avh-demo} in the Appendix.

\paragraph{Clinical Interview Dataset} This set contains semi-structured clinical interviews of San Francisco Bay Area male outpatients diagnosed with SSDs participating in a study of oxytocin conducted independently at University of California, San Francisco (UCSF) \citep{Bradley2024}. All participants are provided with written informed consent and study protocols were approved by the IRB at the UCSF. Following a prior work \citep{poole2000functional}, the clinical assessments were conducted by trained raters in the form of a composite score  combining the conceptual disorganization item (ranging from 1-7 with increasing severity) \citep{Kay1987} from the Positive and Negative Syndrome Scale (PANSS) and the incoherent speech item (ranging from 0-5 with increasing severity) from the Comprehensive Assessment of Symptom and History (CASH) \citep{Andreasen1992} to supplement the disorganized symptom subscale and the measure of suicidality. To avoid any potential confusion between these terms referring to different types of ratings, in the remainder of this paper we will refer this score as composite PANSS. We use manually transcribed interviews from 39 participants with corresponding composite PANSS between 2 and 8 (in the range of 0-12), with a mean of 3.36 and a standard deviation of 1.80. The transcript-level demographic information for this dataset is summarized in Table~\ref{tab:ellen-demo} in Appendix.

\subsection{Language models}
Pythia is the first LLM suite deliberately designed to enable scientific research on LLMs. The Pythia suite offers pre-trained decoder-only autoregressive LMs ranging from 70M to 12B parameters. The Pythia suite is trained on the Pile corpus \citep{gao2020pile}, which is a publicly available and curated collection of English language. In particular, we select Pythia checkpoints (70m, 160m, 410m, 1b, 1.4b, 2.8b, 6.9b, and 12b in parameter size) that are pre-trained on a  \textit{deduplicated} Pile corpus containing approximately 207B tokens. We select these checkpoints as deduplication has demonstrated its benefits in LLM training process \citep{lee-etal-2022-deduplicating}. The Pythia suite largely follows the architecture and hyperparameters of GPT-3, but differs in several aspects: a) it uses fully dense attention layers; b) it is pre-trained using Flash Attention \citep{dao2022flashattention} for improved device throughput; and c) it uses rotary positional embeddings \citep{su2024roformer} for a flexible mechanism to encode positional information.

We also compare Pythia suite with locally-hosted LLaMA-3.1-405b \citep{dubey2024llama} model, which is quantized with 4-bit precision using ExLlamaV2\footnote{\url{https://github.com/turboderp-org/exllamav2}} (prior work indicates that quantization does not significantly degrade model performance \citep{lee2024comprehensive}). As initial experimentation in previous work showed comparable results to those obtained with the base model (without instruction tuning), we use an instruction-tuned model, hosted locally on a secure server.

\subsection{Global and sliding window PPLs}

We compute PPL for a transcript using two approaches: a) a global PPL that evaluates the full transcript as a single sequence; and b) a local PPL using sliding windows of 8, 16, 32, 64, and 128 for the Pythia suite, and a sliding window of 64 for the LLaMA model, as prior work indicates that restricting to a short input (e.g., context length of 128) can substantially improve the performance of LMs \citep{press-etal-2021-shortformer}. The sliding window is defined as a window of a corresponding number of tokens moved sequentially through the transcript. PPL is calculated for each window position as it shifts one token at a time until reaching the end of the transcript. If the transcript is shorter than the designated sliding window, then we calculate the global PPL for the transcript instead. As window size increases, the sliding window PPL approach allows the model to have more dynamic context when making each prediction, resulting in a more accurate approximation of the fully-factorized PPL (i.e., the global PPL). This can be particularly useful for evaluating spontaneous speech where the context is more fragmented than with read speech \citep{auer2009line, shriberg2001errrr, agmon2023s, wang2010breath}. To generate a transcript-level measure, we use the maximum and the averaged PPL across the estimated sliding window PPLs, in addition to the global PPL for each transcript. For each measure, we compare the Spearman $\rho$ with the TALD and composite PANSS scores for the AVH and clinical interview datasets respectively.

We opt to use maximum sliding window PPL as our primary transcript-level metric for detecting incoherent language. The rationale for this choice is evident in the distinct separation between transcripts that exhibit \textit{mild} derailment (with TALD derailment $<$ 3, labeled as 0) and those that exhibit \textit{severe} derailment (with TALD derailment $\geq$ 3, labeled as 1 ) in the AVH dataset (Figure~\ref{fig:avh-max-index} in Appendix). Transcripts rated with TALD $\geq$ 3 consistently exhibit higher maximum sliding window PPL spikes across different model sizes (particularly visible with the sliding window length of 64), while transcripts rated below this threshold maintain relatively stable, lower PPL patterns. We observed a similar trend in Figure~\ref{fig:ellen-max-index} in Appendix, the variation of PPL spikes across different severities of composite PANSS, suggesting that maximum sliding window PPL reflects disorganized speech.\footnote{Our experiments are conducted on 3 H100 GPUs.} 

\section{Results}
\subsection{Global PPL as a proxy for FTD-related linguistic patterns}

\begin{figure*}[htbp]
\centering
\subfigure[Global PPLs of transcripts in the AVH dataset]{
  \includegraphics[width=0.48\textwidth]{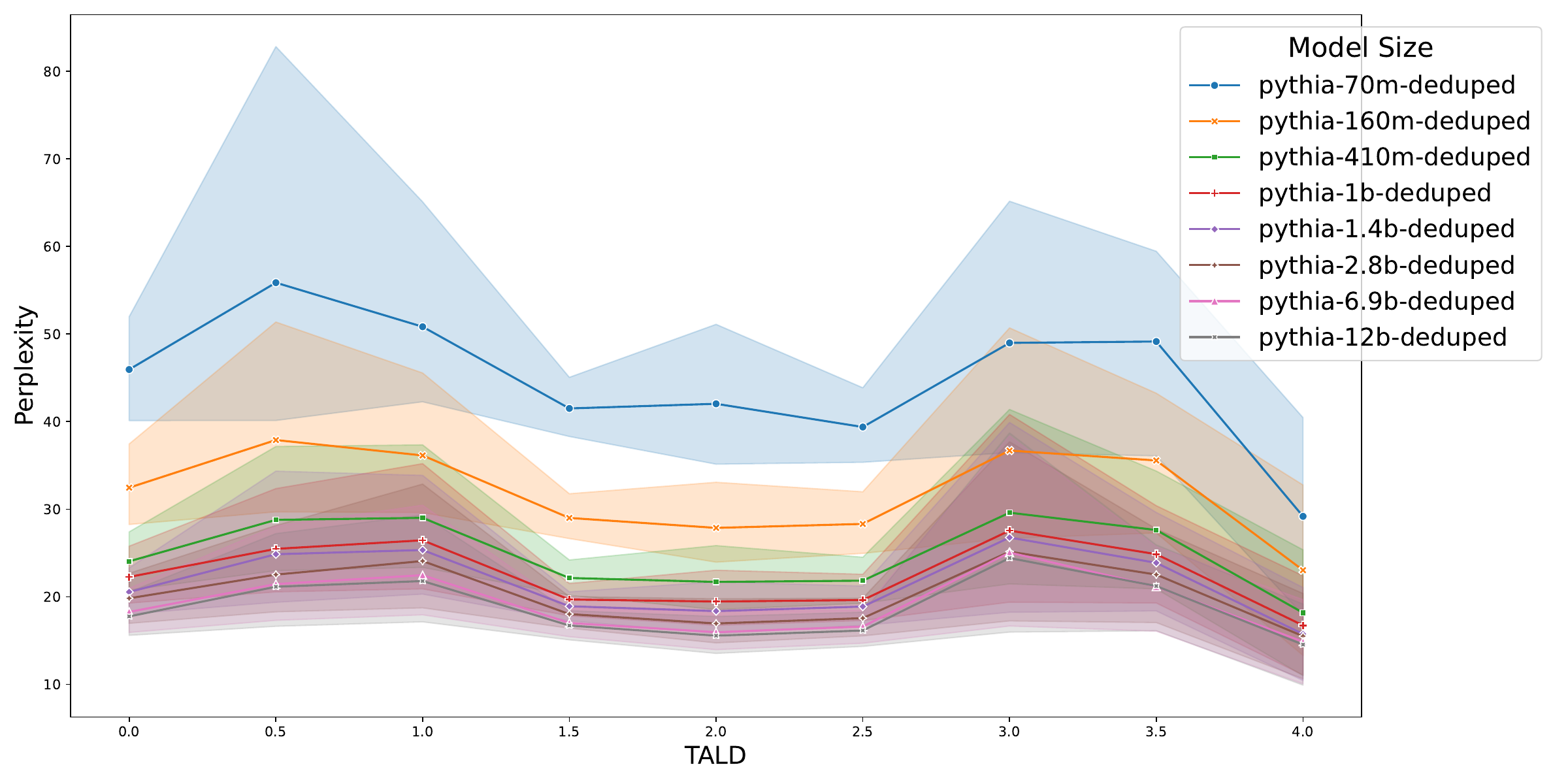}
  \label{fig:global-avh}
}
\hfill
\subfigure[Global PPLs of transcripts in the clinical interview dataset]{
  \includegraphics[width=0.48\textwidth]{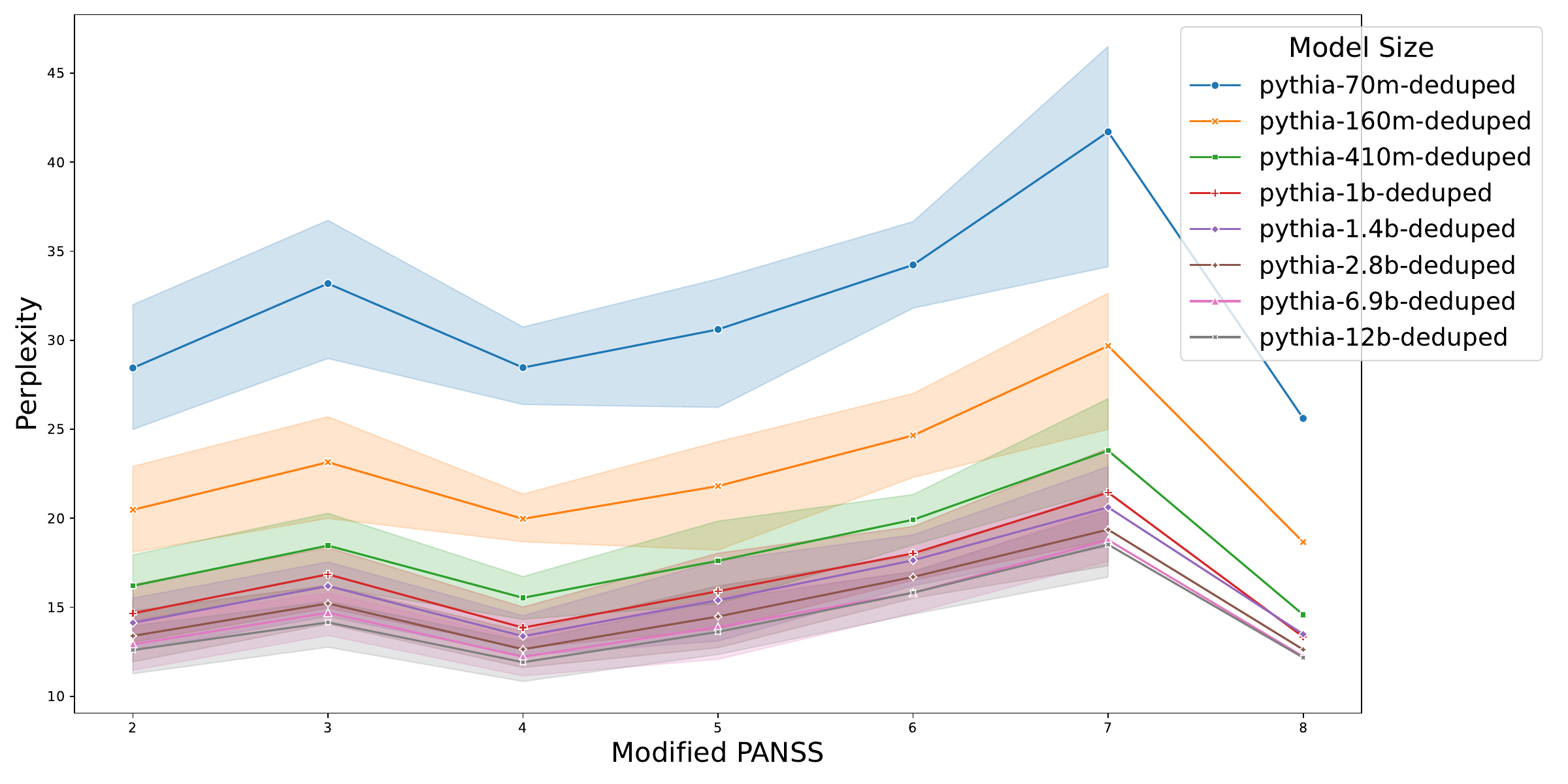}
  \label{fig:global-ellen}
}
\caption{Global PPLs estimated by the Pythia suite. The shaded area represents 95\% confidence intervals of the global PPLs estimated from a pre-trained Pythia model.}
\label{fig:global-ppl}
\end{figure*}

As illustrated in Figure~\ref{fig:global-ppl}, smaller Pythia models (Pythia-70m and Pythia-160m) consistently exhibited higher global PPLs compared to their larger counterparts across both the AVH and clinical interview datasets. Larger models (6.9b and 12b parameters) tended to cluster together at lower PPLs, suggesting diminishing effects on PPL estimation as model size increases. We observed minimal correlation ($\rho<$ 0.01) between global PPL and TALD (in the AVH set),  and this was statistically insignificant across all model sizes. While correlations between global PPL and composite PANSS (in the clinical interview set) were present (Spearman's $\rho$ between 0.20 and 0.39), statistical significance was achieved only for the larger models, including Pythia-2.8b, Pythia-6.9b, and Pythia-12b, at $\alpha=0.1$.

\subsection{Sliding window PPL performance}

\subsubsection{The AVH dataset}

\begin{table}[htbp]
\centering
\resizebox{\columnwidth}{!}{%
\begin{tabular}{@{}c|ccccc@{}}
\toprule
\multirow{2}{*}{\textbf{Model}} & \multicolumn{5}{c}{\textbf{Sliding windows}} \\ \cmidrule(l){2-6} 
 & \multicolumn{1}{c|}{8} & \multicolumn{1}{c|}{16} & \multicolumn{1}{c|}{32} & \multicolumn{1}{c|}{64} & 128 \\ \midrule
70m & \multicolumn{1}{c|}{0.366***} & \multicolumn{1}{c|}{0.375***} & \multicolumn{1}{c|}{0.427***} & \multicolumn{1}{c|}{\textbf{0.440***}} & 0.370*** \\ \midrule
160m & \multicolumn{1}{c|}{0.369***} & \multicolumn{1}{c|}{0.360***} & \multicolumn{1}{c|}{0.426***} & \multicolumn{1}{c|}{\textbf{0.451***}} &  0.378***\\ \midrule
410m & \multicolumn{1}{c|}{0.347***} & \multicolumn{1}{c|}{0.336***} & \multicolumn{1}{c|}{0.430***} & \multicolumn{1}{c|}{\textbf{0.458***}} & 0.378*** \\ \midrule
1b & \multicolumn{1}{c|}{0.329***} & \multicolumn{1}{c|}{0.328***} & \multicolumn{1}{c|}{0.431***} & \multicolumn{1}{c|}{\textbf{0.458***}} & 0.367*** \\ \midrule
1.4b & \multicolumn{1}{c|}{0.331***} & \multicolumn{1}{c|}{0.305***} & \multicolumn{1}{c|}{0.423***} & \multicolumn{1}{c|}{\textbf{0.486***}} & 0.388*** \\ \midrule
2.8b & \multicolumn{1}{c|}{0.315***} & \multicolumn{1}{c|}{0.316***} & \multicolumn{1}{c|}{0.435***} & \multicolumn{1}{c|}{\textbf{0.464***}} & 0.365*** \\ \midrule
6.9b & \multicolumn{1}{c|}{0.319***} & \multicolumn{1}{c|}{0.310***} & \multicolumn{1}{c|}{0.420***} & \multicolumn{1}{c|}{\textbf{0.475***}} & 0.370*** \\ \midrule
12b & \multicolumn{1}{c|}{0.317***} & \multicolumn{1}{c|}{0.307***} & \multicolumn{1}{c|}{0.421***} & \multicolumn{1}{c|}{\textbf{0.470***}} & 0.372*** \\ \midrule
LLaMA & \multicolumn{1}{c|}{--} & \multicolumn{1}{c|}{--} & \multicolumn{1}{c|}{--} & \multicolumn{1}{c|}{0.457***} & -- \\ \bottomrule
 \addlinespace[1ex]
\multicolumn{3}{l}{\textsuperscript{***}$p<0.01$, 
  \textsuperscript{**}$p<0.05$, 
  \textsuperscript{*}$p<0.1$}
\end{tabular}%
}
\caption{The AVH dataset Spearman's $\rho$ between the \textit{maximum} sliding window PPL and TALD across model size. \textbf{Bold} indicates the highest $\rho$ for a model.}
\label{tab:avh-max}
\end{table}

As shown in Table~\ref{tab:avh-max}, all correlations between \textit{maximum} sliding window PPL and TALD scores were statistically significant (p-value $<$ 0.01) across all model and sliding window sizes. The strongest correlations consistently occurred with a 64-token sliding window, with coefficients peaking at 0.486 for the 1.4b model, and remaining moderately correlated with TALD scores for all model variants. Interestingly, the 4-bit quantized LLaMA 405b model did not outperform the Pythia suite, attaining a lower Spearman $\rho$ of 0.457 on the 64-token sliding window.

%While the degree of correlations did not show a clear linear relationship with model size, as both smaller (70m and 160m) and larger models (6.9b and 12b) maintained comparable correlation coefficients within each window size, we observed that the correlation coefficients began to decline slightly after the model reaching the 410m parameter size for the sliding windows of 8, 16, 32, and 128. This decline in the 64 sliding window began after the model reaching the 1.4b parameter size, which collectively suggests a potential plateau or diminishing returns in correlation with increased model size.

Table~\ref{tab:avh-avg} in Appendix shows similar patterns of correlation between  \textit{averaged} sliding window PPL for a transcript and TALD scores. While all correlations remain significant (p-value $<$ 0.01) across all model sizes and sliding windows, the correlation coefficients are generally lower compared to those for maximum sliding window PPLs. The 64-token sliding window again emerged as the optimal configuration, with correlation coefficients ranging from 0.202 to 0.251. The 1.4b model achieved the strongest correlation ($\rho=$ 0.251), followed closely by the 6.9b model ($\rho=$ 0.249). The LLaMA model achieved the highest correlation ($\rho=$ 0.371) for the sliding window of 64 tokens.

%While there is a modest trend of increasing correlation with model size, this improvement plateaued and slightly decreased after reaching the 410m parameter size for the sliding windows of 8, 16, 32, and 128. We observed a similar trend for the 64 sliding window, where the correlation peaked when the model size reached 6.9b, and declined at 12b.

\subsubsection{The clinical interview dataset}

\begin{table}[htbp]
\centering
\resizebox{\columnwidth}{!}{%
\begin{tabular}{@{}c|ccccc@{}}
\toprule
\multirow{2}{*}{\textbf{Model}} & \multicolumn{5}{c}{\textbf{Sliding windows}} \\ \cmidrule(l){2-6} 
 & \multicolumn{1}{c|}{8} & \multicolumn{1}{c|}{16} & \multicolumn{1}{c|}{32} & \multicolumn{1}{c|}{64} & 128 \\ \midrule
70m & \multicolumn{1}{c|}{0.265} & \multicolumn{1}{c|}{\textbf{0.482***}} & \multicolumn{1}{c|}{0.338**} & \multicolumn{1}{c|}{0.356**} & 0.344** \\ \midrule
160m & \multicolumn{1}{c|}{0.414***} & \multicolumn{1}{c|}{0.433***} & \multicolumn{1}{c|}{0.316*} & \multicolumn{1}{c|}{0.354***} &  0.325**\\ \midrule
410m & \multicolumn{1}{c|}{0.385*} & \multicolumn{1}{c|}{0.433***} & \multicolumn{1}{c|}{0.316*} & \multicolumn{1}{c|}{0.354**} & 0.325** \\ \midrule
1b & \multicolumn{1}{c|}{0.415***} & \multicolumn{1}{c|}{0.352**} & \multicolumn{1}{c|}{0.380**} & \multicolumn{1}{c|}{0.410***} & 0.313* \\ \midrule
1.4b & \multicolumn{1}{c|}{0.458***} & \multicolumn{1}{c|}{0.370***} & \multicolumn{1}{c|}{0.382**} & \multicolumn{1}{c|}{0.418***} & 0.326** \\ \midrule
2.8b & \multicolumn{1}{c|}{0.425***} & \multicolumn{1}{c|}{0.385**} & \multicolumn{1}{c|}{0.369**} & \multicolumn{1}{c|}{\textbf{0.428***}} & 0.348*** \\ \midrule
6.9b & \multicolumn{1}{c|}{\textbf{0.478***}} & \multicolumn{1}{c|}{0.352**} & \multicolumn{1}{c|}{0.394**} & \multicolumn{1}{c|}{0.414***} & \textbf{0.368**} \\ \midrule
12b & \multicolumn{1}{c|}{0.441***} & \multicolumn{1}{c|}{0.313*} & \multicolumn{1}{c|}{\textbf{0.404**}} & \multicolumn{1}{c|}{0.412***} & 0.357* \\ \midrule
LLaMA & \multicolumn{1}{c|}{--} & \multicolumn{1}{c|}{--} & \multicolumn{1}{c|}{--} & \multicolumn{1}{c|}{0.249} & -- \\ \bottomrule
 \addlinespace[1ex]
\multicolumn{3}{l}{\textsuperscript{***}$p<0.01$, 
  \textsuperscript{**}$p<0.05$, 
  \textsuperscript{*}$p<0.1$}
\end{tabular}%
}
\caption{The clinical interview dataset Spearman's $\rho$ between the \textit{maximum} sliding window PPL and modified PANSS across model size. \textbf{Bold} indicates the highest $\rho$ for a model.}
\label{tab:ellen-max}
\end{table}

The patterns observed in Table~\ref{tab:ellen-max} of the \textit{maximum} sliding window PPLs on clinical interview dataset vary compared to those from audio diary data shown in Table~\ref{tab:avh-max}. Pythia-70m with a sliding window of 16 has the highest correlation ($\rho=$ 0.482, p-value $<$ 0.01), while the Pythia-6.9b model shows comparable performance with a sliding window of 8 ($\rho=$ 0.478, p-value $<$ 0.01). Unlike the patterns with TALD, the sliding window size of 64 tokens was not optimal across all models, though it did yield strong correlations for several model sizes, particularly with the Pythia-2.8b model ($\rho=$ 0.428, p-value $<$ 0.01). Similarly, we also observed that LLaMA achieved the lowest and insignificant Spearman $\rho$ with a 64-token sliding window.

Table~\ref{tab:ellen-avg} in Appendix shows more moderate relationships for \textit{averaged} sliding window PPLs in the clinical interview data, with correlation coefficients ranging from 0.248 to 0.360. The Pythia-1.4b model demonstrated consistently strong performance across all window sizes, achieving the highest correlation coefficients of all models, with its peak at a sliding window of 64 ($\rho=$ 0.360, p-value $<$ 0.05). LLaMA achieved the lowest Spearman $\rho$ in the sliding window of 64, despite this being moderately significant.

\subsection{The comparison of model sizes and sliding windows}

For TALD correlations (Table~\ref{tab:avh-max} and Table~\ref{tab:avh-avg} in Appendix), larger sliding window sizes (32 and 64) consistently showed stronger correlation as model size increased. However, this trend was less evident with smaller sliding window sizes (8 and 16), where the correlation coefficients remained relatively stable across model sizes. In contrast, the composite PANSS correlations (Table~\ref{tab:ellen-max} and Table~\ref{tab:ellen-avg} in Appendix) exhibited a different pattern: smaller sliding window sizes (8 and 16) showed more variation across model sizes with maximum PPLs, with correlation coefficients fluctuating. For example, with the sliding window of 8, the correlation coefficients ranged from $\rho=$ 0.265 to $\rho=$ 0.478. The correlation coefficients with averaged PPLs show more consistent behavior, as they gradually increase up to the 1.4b model across all sliding window sizes before they plateau or slightly decline with in larger models. %This indicates that larger model size does not necessarily translate to stronger correlation with clinical scales. 

As can be observed in Table~\ref{tab:avh-max}, which shows the correlations between maximum PPL for a transcript and the TALD on the AVH dataset, all models exhibit a consistent pattern where correlation coefficients generally increase with a sliding window of 8 and 16, peak at a sliding window of 64, and then decrease with a sliding window of 128. There is a similar trend in Table~\ref{tab:avh-avg}, but with more moderate increases and decreases. In contrast, there are more variable patterns with the clinical interview results shown in Table~\ref{tab:ellen-max}. Smaller models (70m-410m) tended to achieve peak correlations at smaller window sizes, while larger models show more distributed peaks across different window sizes. The correlation between the averaged sliding window PPL and composite PANSS shows the most consistent pattern across window sizes, as is particularly evident with the 1.4b model, which maintains relatively stable correlations ranging from $\rho=$ 0.272 to $\rho=$ 0.360 across all window sizes. Notably, a sliding window of 128 consistently produced the weakest correlations in both datasets, suggesting that larger window size may dampen the PPL response to local patterns as compared with medium-sized windows. Interestingly, there is also a general trend of diminishing effects of sliding window size with both datasets, with the correlation coefficients declining with larger models (e.g., billions of parameter size) at the same sliding window size.
%suggesting that larger sliding window sizes may provide more robust, though not necessarily stronger, correlations regardless of model size.

\section{Discussion}
 
Our key findings are as follows. First, our results suggest LM PPL can potentially serve as an objective computational marker for capturing subtle linguistic patterns associated with FTD. This aligns with previous studies indicating that such abnormal linguistic patterns manifest in ways that can be quantitatively measured \citep{colla2022semantic, he2024navigating, xu2021centroid, XU2022103998, sharpe2024gpt}. Second, our results extend \citet{sharpe2024gpt}'s work by examining fine-grained sliding window PPLs to capture semantic variations across longer sequences using models ranging from 70m to 405b parameters for two model families. Our results indicate that model size does not necessarily correlate linearly with its capability for detecting FTD-related linguistic manifestations. Furthermore, our findings suggest that small/medium-sized sliding windows consistently demonstrate optimal performance across different model sizes, indicating an effective balance point between clinical utility and computational efficiency. That performance declines with larger windows may suggest the approximate range within which contextual inconsistencies manifest in FTD. These findings collectively suggest that calibrated smaller LMs can be at least as effective as their larger counterparts, offering practical advantages for real-world deployment while maintaining clinically-validated and robust performance in detecting FTD-related language differences.

Our work also demonstrates a more nuanced relationship between window size and the linguistic manifestations of FTD. While prior work \citep{sharpe2024gpt} using GPT-3 indicates that differences in lexical probability (i.e., the intermediate products of PPLs) differ more between cases and controls with larger context windows (i.e., up to 50 tokens) for FTD, our work provides a more granular characterization of optimal sliding window sizes for alignment with human ratings in both monologue and conversational speech samples. This is particularly important for detecting linguistic inconsistencies that span across longer text segments, an aspect of comprehension that prior research suggests may be selectively impaired in people with SSDs \citep{kuperberg2010language1,kuperberg2010language2,sharpe2024gpt}. In addition, it reveals diminishing correlation coefficients at larger model and sliding window size, suggesting an upper bound to the utility of increasing both context window size and model size. Interestingly, the clinical interview dataset shows more variable optimal sliding window sizes across different model scales, with smaller models performing best at shorter windows and larger models showing distributed peaks across different window sizes. These varied patterns suggest that the manifestation of FTD -- particularly in conversational language -- may operate at multiple scales, rather than simply becoming more apparent with larger contexts. This in turn suggests that such LM-based methods may require calibrated combinations of both model size and context window, rather than simply maximizing either dimension.

Our findings demonstrate the generalizability of PPL-based computational and automated assessment across both monologue (AVH) and conversational data (clinical interview), suggesting that changes in language associated with FTD can be effectively captured regardless of the communicative setting. Spontaneous speech presents unique complexities due to its impromptu nature, where speakers have minimal time to organize their thoughts. The challenges include a lack of clear syntactic boundaries \citep{auer2009line, shriberg2001errrr, agmon2023s, wang2010breath}, complex interaction of linguistic demands due to mental states \citep{menn1989cross, CAPLAN1998184}, and context sensitivity. They make it particularly challenging for generalizable computational analysis. However, our results show that PPL-based measures can effectively operate within these complexities, yielding statistically significant correlations with human ratings across both monologue and conversational datasets. This capability to perform consistently across different communication contexts is important for clinical applications, where assessment tools may need to maintain reliability across various real-world scenarios. The consistency of our results across both data sources indicates the potential of sliding-window based LM perplexity as an automated and computational assessment tool.

With respect to model size, 4-bit quantized LLaMA 405b, despite its significantly larger scale and strong performance on open-domain tasks \citep{lee2024comprehensive}, consistently underperformed compared to smaller Pythia models. This finding supports our hypothesis regarding the potential advantages of smaller LMs in detecting subtle linguistic patterns associated with FTD, though it remains to be determined whether this advantage is attributable to model size or a relatively constrained amount of training data. Larger LMs, with their extensive pre-training on vast corpora of text of uncertain provenance, may find the subtle linguistic patterns that characterize FTD more predictable. In contrast, smaller LMs' more limited exposure to coherent language patterns and/or constrained capacity (as proxied by parameter size) may paradoxically enhance their sensitivity to linguistic patterns associated with FTD. This suggests that the relationship between model scale and clinical assessment capability is not strictly linear (i.e., bigger is not necessarily better), and that optimal performance may be achieved by models that maintain adequate linguistic competence while remaining sensitive to deviations from typical language patterns. A further advantage of the Pythia suite is that their training data are publicly available, and therefore amenable to analyses to identify sources of biased assessment, such as the absence of training data reflecting dialectical variation characteristic of particular population groups. These findings and observations collectively suggest that pursuing ever-larger models may not necessarily yield better clinical assessment capabilities and utilities.

Our analysis across multiple model sizes provides empirical guidance for sliding window size selection in clinical practice. The finding that small- to medium-sized sliding windows (typically 16 to 64 tokens) consistently demonstrate optimal performance across different model sizes suggests an effective range for practical implementation. This observation is consistent with previous studies demonstrating that linguistic inconsistencies manifest as local coherence disruptions \citep{sitnikova2002electrophysiological, swaab2013spared,kuperberg2010language1, kuperberg2010language2}. The observed performance decline with larger windows ($>$ 128 tokens) further supports this understanding. Notably, the optimal sliding window sizes remain consistent across both shorter monologues ($\approx$ 100 tokens) and longer clinical interviews ($>$ 1000 tokens), suggesting that the linguistic manifestation of FTD operates at a fragmented level independent of overall discourse length or interaction type. This pattern suggests that aspects of FTD may be best characterized in intermittent steps rather than as global narrative incoherence. Additionally, the sliding window size sensitivity remains remarkably consistent across model scales from 70m to 12b, suggesting that PPL, as a computational marker, can effectively capture such linguistic manifestations, providing a compelling evidence for the context-sensitive nature of FTD and its variable expression across different communicative demands.

Our findings suggest that PPL derived from smaller LMs with granular sliding windows offer promising clinical utility in addition to existing assessment methods. Furthermore, The reduced computational requirements of our approach also makes it particularly suitable for resource-constrained settings, potentially enabling automated FTD screening in underserved communities. These models' ability to detect subtle linguistic manifestations of FTD opens several promising application pathways in clinical practice. Most notably, the efficiency of smaller LMs (70m-410m parameters) enables privacy-preserving, on-device processing that could streamline the mental health monitoring and early intervention. For example, these lightweight models could be integrated into telehealth platforms to analyze discourse during remote psychiatric consultations in real-time, providing clinicians with immediate linguistic computational markers while ensuring all patient data remains on local devices. In ambient monitoring scenarios, these models could be deployed on smartphones to periodically assess everyday conversation with participants' prior consent, creating longitudinal datasets that track subtle changes in FTD, enabling the comparison of cross-sectional linguistic patterns to identify preliminary warning signs that might surf unnoticed. The resulting data could also used for near-real-time flagging of warning signs, which could be shared with their clinical teams to enable time-sensitive interventions that may help prevent further deterioration \citep{ben2017crosscheck}.

While our token-level sliding window PPL method demonstrates promising results, we acknowledge that certain sentence-level proximity-based methods \citep{XU2022103998} achieved comparable or higher correlations. However, the sliding window PPL method is responsive to linguistic inconsistencies at varying granularities that complement existing methods, potentially capturing dynamic aspects of FTD that may be missed by static sentence-level measures.

\section{Conclusion}

We presented experimental findings consistent with prior work suggesting sliding-window PPL as an efficient measurement for linguistic patterns associated with FTD. Surprisingly, our findings suggest that smaller LMs with calibrated sliding window sizes, are more sensitive to such linguistic manifestations. The comparable effectiveness of smaller models opens new possibilities for implementing automated and computational language assessment tools in resource-constrained clinical settings while remaining cost-efficient and privacy-preserving.

\newpage
\section*{Limitations}

The work presented here has several limitations. All participants represented in both data sets are English speakers, and it remains unclear the extent to which our findings apply to other languages. Our analysis relied on transcribed speech data, which may not fully capture the nuances of spoken language, including prosody, pauses, and other paralinguistic features that could be clinically relevant (for a related review, see \citet{ehlenLinguisticFindingsPersons2023}). While our findings demonstrate correlations between PPL and human ratings, these ratings do not constitute clinical diagnoses, which would be needed for case/control comparisons. Furthermore our analysis does not account for potential confounding variables -- such as age, gender, origin and socioeconomic status -- which may influence language patterns. While smaller LMs show promising results, we have not yet established clear clinical thresholds that would be necessary for diagnosis, or assessed the utility of measurements over time as indicators of change in clinical status. We also note that the severity scores for both datasets are relatively low on average, and that datasets with more representation of severe FTD may be needed to establish optimal parameter settings in this context. Our study focused specifically on positive FTD, a key diagnostic feature for SSDs. Therefore, the extent to which sliding window PPL is responsive to linguistic manifestations of other conditions remains to be established. Future work to address these limitations will be required to reach the potential of these methods for clinical deployment. Finally, while we included a larger language model with a 64-token sliding window, including additional sliding window sizes with LLaMA would make for a more comprehensive analysis that is left for future work.

\section*{Acknowledgment}

This work was supported by National Institute of Mental Health under award number R01MH112641, U01MH13590, and the Department of Veterans Affairs under award number IK1 CX002092-02. We also want to thank the reviewers for their insightful critiques and suggestions that helped improve this manuscript.

\bibliography{anthology,custom}
%\bibliography{custom}

\appendix

\section{Appendix}
\setcounter{figure}{0}
\renewcommand\thefigure{A.\arabic{figure}}

\setcounter{table}{0}
\renewcommand\thetable{A.\arabic{table}}

\begin{table*}
\centering
\resizebox{0.8\linewidth}{!}{%
\begin{tabular}[t]{llll}
\toprule
  & Level & Mild & Severe\\
\midrule
\# of transcripts &  & 292 & 18\\
Age (mean (SD)) &  & 40.45 (10.71) & 38.33 (8.97)\\
Gender (\%) & Female & 162 (55.5) & 10 (55.6)\\
 & Male & 118 (40.4) & 8 (44.4)\\
 & Transgendered: FTM & 3 (1.0) & 0 (0.0)\\
\addlinespace
 & Transgendered: MTF & 9 (3.1) & 0 (0.0)\\
Education (\%) &  & 1 (0.3) & 0 (0.0)\\
 & Associates Degree & 51 (17.5) & 3 (16.7)\\
 & Bachelors Degree & 25 (8.6) & 0 (0.0)\\
 & Doctorate Degree & 3 (1.0) & 0 (0.0)\\
\addlinespace
 & Grade school & 4 (1.4) & 4 (22.2)\\
 & High School Diploma /GED & 171 (58.6) & 9 (50.0)\\
 & Junior High & 22 (7.5) & 2 (11.1)\\
 & Masters Degree & 15 (5.1) & 0 (0.0)\\
Race (\%) &  & 1 (0.3) & 0 (0.0)\\
\addlinespace
 & American Indian or Alaskan Native & 6 (2.1) & 0 (0.0)\\
 & Asian & 6 (2.1) & 0 (0.0)\\
 & Black or African American & 61 (20.9) & 8 (44.4)\\
 & More than one race & 35 (12.0) & 2 (11.1)\\
 & White & 183 (62.7) & 8 (44.4)\\
\addlinespace
\# of words per transcript (mean (SD)) &  & 182.76 (139.17) & 300.50 (170.42)\\
TALD (mean (SD)) &  & 1.08 (0.70) & 3.33 (0.34)\\
\bottomrule
\end{tabular}
}
\caption{Basic transcript-level demographic information for the AVH dataset. \texttt{Mild} denotes as mild symptoms of positive FTD where TALD score $<$ 3, and \texttt{Severe} denotes severe symptoms of positive FTD, where TALD score $\geq$ 3.}
\label{tab:avh-demo}
\end{table*}

\begin{table*}
\centering
\resizebox{0.9\linewidth}{!}{%
\begin{tabular}[t]{lllllllll}
\toprule
  & level & 2 & 3 & 4 & 5 & 6 & 7 & 8\\
\midrule
\# of transcripts &  & 20 & 5 & 5 & 3 & 2 & 3 & 1\\
Gender (\%) & Male & 20 (100.0) & 5 (100.0) & 5 (100.0) & 3 (100.0) & 2 (100.0) & 3 (100.0) & 1 (100.0)\\
Age (mean (SD)) &  & 32.35 (10.84) & 28.20 (6.26) & 35.20 (13.77) & 41.00 (3.61) & 24.50 (3.54) & 36.00 (23.39) & 58.00 (NA)\\
Race (\%) & African American & 4 (20.0) & 0 (0.0) & 2 (40.0) & 0 (0.0) & 0 (0.0) & 1 (33.3) & 0 (0.0)\\
 & Asian & 2 (10.0) & 1 (20.0) & 0 (0.0) & 0 (0.0) & 0 (0.0) & 0 (0.0) & 1 (100.0)\\
\addlinespace
 & Latino & 1 (5.0) & 1 (20.0) & 0 (0.0) & 0 (0.0) & 1 (50.0) & 0 (0.0) & 0 (0.0)\\
 & Other (including mixed) & 2 (10.0) & 0 (0.0) & 1 (20.0) & 1 (33.3) & 1 (50.0) & 1 (33.3) & 0 (0.0)\\
 & White & 11 (55.0) & 3 (60.0) & 2 (40.0) & 2 (66.7) & 0 (0.0) & 1 (33.3) & 0 (0.0)\\
Education (\%) & Associates Degree & 2 (10.0) & 0 (0.0) & 0 (0.0) & 0 (0.0) & 0 (0.0) & 1 (33.3) & 0 (0.0)\\
 & Bachelors Degree & 3 (15.0) & 2 (40.0) & 2 (40.0) & 1 (33.3) & 0 (0.0) & 0 (0.0) & 0 (0.0)\\
\addlinespace
 & GED & 1 (5.0) & 1 (20.0) & 0 (0.0) & 1 (33.3) & 0 (0.0) & 0 (0.0) & 0 (0.0)\\
 & H.S. Diploma & 13 (65.0) & 1 (20.0) & 3 (60.0) & 1 (33.3) & 2 (100.0) & 2 (66.7) & 1 (100.0)\\
 & Other & 0 (0.0) & 1 (20.0) & 0 (0.0) & 0 (0.0) & 0 (0.0) & 0 (0.0) & 0 (0.0)\\
 & Vocational Certification & 1 (5.0) & 0 (0.0) & 0 (0.0) & 0 (0.0) & 0 (0.0) & 0 (0.0) & 0 (0.0)\\
\# of words per transcript (mean (SD)) &  & 1494.20 (1053.27) & 1178.40 (536.35) & 1822.00 (1774.69) & 1478.67 (601.23) & 2004.50 (340.12) & 3023.67 (2197.36) & 4339.00 (NA)\\
\bottomrule
\end{tabular}
}
\caption{Basic transcript-level demographic information for the clinical interview dataset. The top row represents the value of the composite PANSS.}
\label{tab:ellen-demo}
\end{table*}

\begin{figure*}[htbp]
    \centering
    \includegraphics[width=0.8\textwidth]{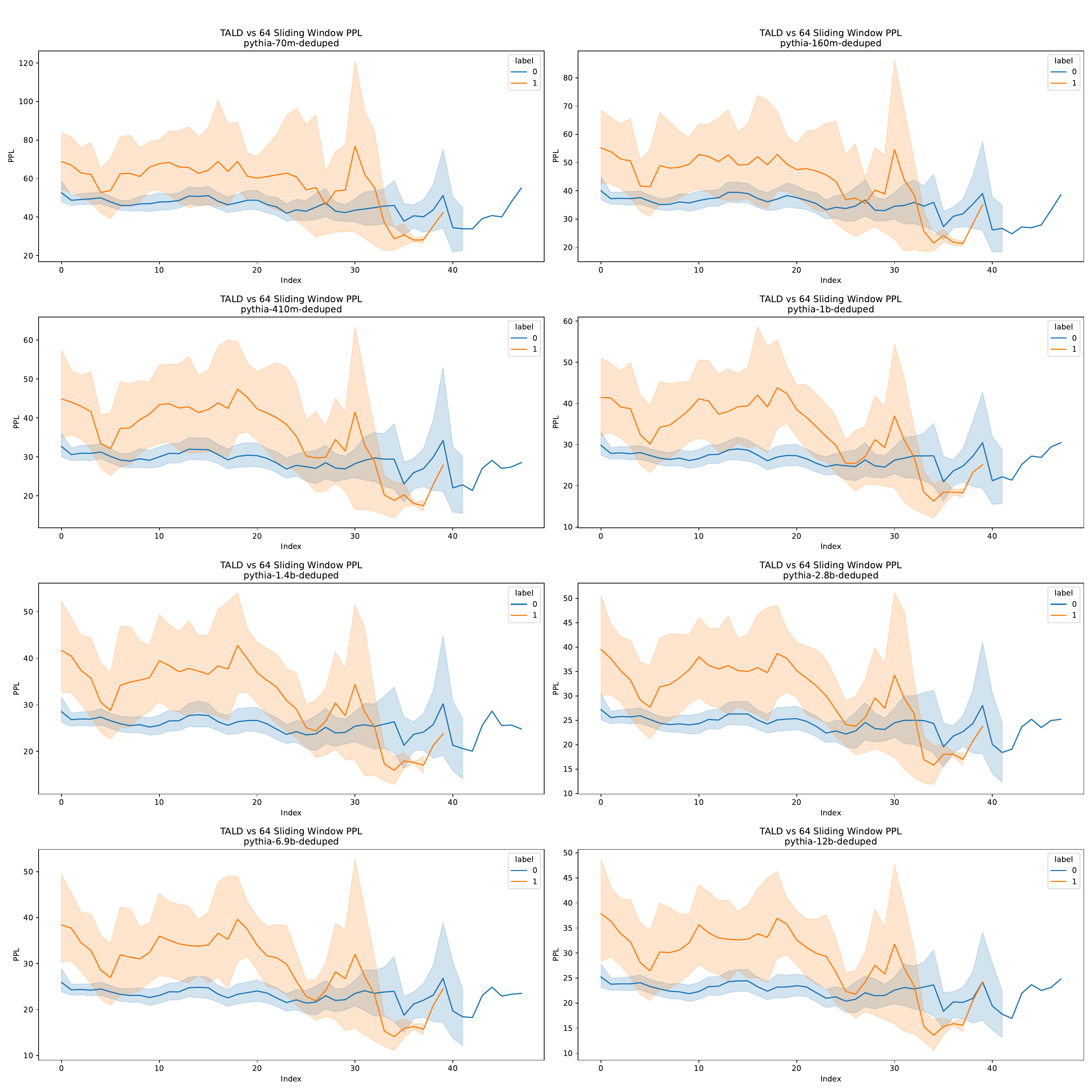}
    \caption{The distribution of sliding window PPLs using a sliding window of 64 tokens on the AVH dataset, where the x-axis represents the index of sliding window PPLs in transcripts. The shaded area represents the 95\% confidence interval of the estimated sliding window PPL on a given index. The label of 0 is defined where TALD $<$ 3, serving as a proxy label for cognitively healthy individuals, whereas the label of 1 serves as the proxy label of FTD individuals.}
    \label{fig:avh-max-index}
\end{figure*}

\begin{figure*}[htbp]
    \centering
    \includegraphics[width=0.8\textwidth]{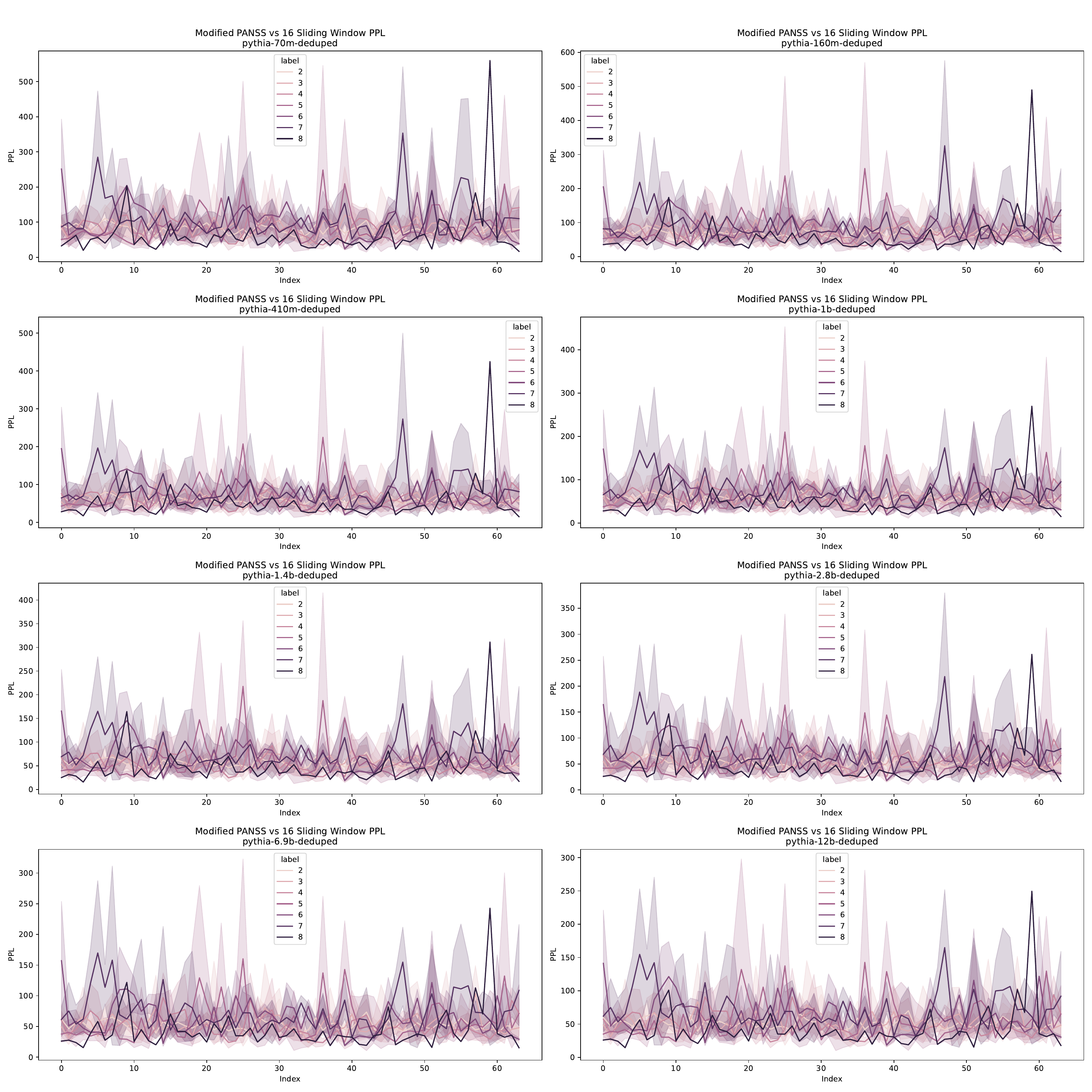}
    \caption{The distribution of sliding window PPLs using the sliding window of 16 tokens on the clinical interview dataset, where the x-axis represents the index of sliding window PPLs in transcripts. The shaded area represents the 95\% confidence interval of the estimated sliding window PPL on a given index.}
    \label{fig:ellen-max-index}
\end{figure*}

\begin{table}[htbp]
\centering
\resizebox{\columnwidth}{!}{%
\begin{tabular}{@{}c|ccccc@{}}
\toprule
\multirow{2}{*}{\textbf{Model}} & \multicolumn{5}{c}{\textbf{Sliding windows}} \\ \cmidrule(l){2-6} 
 & \multicolumn{1}{c|}{8} & \multicolumn{1}{c|}{16} & \multicolumn{1}{c|}{32} & \multicolumn{1}{c|}{64} & 128 \\ \midrule
70m & \multicolumn{1}{c|}{0.158***} & \multicolumn{1}{c|}{0.156***} & \multicolumn{1}{c|}{0.176***} & \multicolumn{1}{c|}{\textbf{0.202***}} & 0.180*** \\ \midrule
160m & \multicolumn{1}{c|}{0.177***} & \multicolumn{1}{c|}{0.170***} & \multicolumn{1}{c|}{0.177***} & \multicolumn{1}{c|}{\textbf{0.206***}} &  0.183***\\ \midrule
410m & \multicolumn{1}{c|}{0.175***} & \multicolumn{1}{c|}{0.179***} & \multicolumn{1}{c|}{0.196***} & \multicolumn{1}{c|}{\textbf{0.225***}} & 0.189*** \\ \midrule
1b & \multicolumn{1}{c|}{0.178***} & \multicolumn{1}{c|}{0.176***} & \multicolumn{1}{c|}{0.205***} & \multicolumn{1}{c|}{\textbf{0.230***}} & 0.180*** \\ \midrule
1.4b & \multicolumn{1}{c|}{0.181***} & \multicolumn{1}{c|}{0.179***} & \multicolumn{1}{c|}{0.208***} & \multicolumn{1}{c|}{\textbf{0.251***}} & 0.201*** \\ \midrule
2.8b & \multicolumn{1}{c|}{0.168***} & \multicolumn{1}{c|}{0.165***} & \multicolumn{1}{c|}{0.204***} & \multicolumn{1}{c|}{\textbf{0.240***}} & 0.194*** \\ \midrule
6.9b & \multicolumn{1}{c|}{0.172***} & \multicolumn{1}{c|}{0.169***} & \multicolumn{1}{c|}{0.206***} & \multicolumn{1}{c|}{\textbf{0.249***}} & 0.194*** \\ \midrule
12b & \multicolumn{1}{c|}{0.175***} & \multicolumn{1}{c|}{0.174***} & \multicolumn{1}{c|}{0.204***} & \multicolumn{1}{c|}{\textbf{0.245***}} & 0.195*** \\ \midrule
LLaMA & \multicolumn{1}{c|}{--} & \multicolumn{1}{c|}{--} & \multicolumn{1}{c|}{--} & \multicolumn{1}{c|}{0.371***} & -- \\ \bottomrule
 \addlinespace[1ex]
\multicolumn{3}{l}{\textsuperscript{***}$p<0.01$, 
  \textsuperscript{**}$p<0.05$, 
  \textsuperscript{*}$p<0.1$}
\end{tabular}%
}
\caption{The AVH dataset Spearman's $\rho$ between the \textit{averaged} sliding window PPL and TALD across model size. \textbf{Bold} indicates the highest $\rho$ for a model.}
\label{tab:avh-avg}
\end{table}

\begin{table}[htbp]
\centering
\resizebox{\columnwidth}{!}{%
\begin{tabular}{@{}c|ccccc@{}}
\toprule
\multirow{2}{*}{\textbf{Model}} & \multicolumn{5}{c}{\textbf{Sliding windows}} \\ \cmidrule(l){2-6} 
 & \multicolumn{1}{c|}{8} & \multicolumn{1}{c|}{16} & \multicolumn{1}{c|}{32} & \multicolumn{1}{c|}{64} & 128 \\ \midrule
70m & \multicolumn{1}{c|}{0.258} & \multicolumn{1}{c|}{0.248} & \multicolumn{1}{c|}{0.274*} & \multicolumn{1}{c|}{0.276*} & 0.276* \\ \midrule
160m & \multicolumn{1}{c|}{0.264} & \multicolumn{1}{c|}{0.278*} & \multicolumn{1}{c|}{0.296*} & \multicolumn{1}{c|}{0.313*} & 0.294*\\ \midrule
410m & \multicolumn{1}{c|}{0.263} & \multicolumn{1}{c|}{0.276*} & \multicolumn{1}{c|}{0.324**} & \multicolumn{1}{c|}{0.318**} & 0.301*\\ \midrule
1b & \multicolumn{1}{c|}{0.266} & \multicolumn{1}{c|}{0.292*} & \multicolumn{1}{c|}{0.318**} & \multicolumn{1}{c|}{0.330***} & 0.305* \\ \midrule
1.4b & \multicolumn{1}{c|}{\textbf{0.272*}} & \multicolumn{1}{c|}{\textbf{0.334**}} & \multicolumn{1}{c|}{\textbf{0.355**}} & \multicolumn{1}{c|}{\textbf{0.360**}} & \textbf{0.342**} \\ \midrule
2.8b & \multicolumn{1}{c|}{0.261} & \multicolumn{1}{c|}{0.324**} & \multicolumn{1}{c|}{0.344**} & \multicolumn{1}{c|}{0.343**} & 0.325** \\ \midrule
6.9b & \multicolumn{1}{c|}{0.269*} & \multicolumn{1}{c|}{0.315*} & \multicolumn{1}{c|}{0.342**} & \multicolumn{1}{c|}{0.315*} & 0.310* \\ \midrule
12b & \multicolumn{1}{c|}{0.270*} & \multicolumn{1}{c|}{0.302*} & \multicolumn{1}{c|}{0.338**} & \multicolumn{1}{c|}{0.334**} & 0.326** \\ \midrule
LLaMA & \multicolumn{1}{c|}{--} & \multicolumn{1}{c|}{--} & \multicolumn{1}{c|}{--} & \multicolumn{1}{c|}{0.200*} & -- \\ \bottomrule
 \addlinespace[1ex]
\multicolumn{3}{l}{\textsuperscript{***}$p<0.01$, 
  \textsuperscript{**}$p<0.05$, 
  \textsuperscript{*}$p<0.1$}
\end{tabular}%
}
\caption{The clinical interview dataset Spearman's $\rho$ between the \textit{averaged} sliding window PPL and composite PANSS across model size. \textbf{Bold} indicates the highest $\rho$ for a model.}
\label{tab:ellen-avg}
\end{table}

\end{document}